\documentclass[10pt,twocolumn,letterpaper]{article}

\usepackage[pagenumbers]{cvpr}

\definecolor{cvprblue}{rgb}{0.21,0.49,0.74}
\usepackage[pagebackref,breaklinks,colorlinks,allcolors=cvprblue]{hyperref}

\usepackage{algorithm} 
\usepackage{algorithmic} 
\usepackage{amsmath} 

\title{Noise Diffusion for Enhancing Semantic Faithfulness in Text-to-Image Synthesis}

\author{
\small Boming Miao\textsuperscript{1}, Chunxiao Li\textsuperscript{1}, Xiaoxiao Wang\textsuperscript{2}, Andi Zhang\textsuperscript{3}, Rui Sun\textsuperscript{4}, Zizhe Wang\textsuperscript{5}, Yao Zhu\textsuperscript{5}\thanks{Corresponding author: Yao Zhu.}  \\
\small \textsuperscript{1}Beijing Normal University, \small \textsuperscript{2}University of Chinese Academy of Sciences,
\small \textsuperscript{3}University of Manchester,\\ \small \textsuperscript{4}The Chinese University of Hong Kong, Shenzhen, \small \textsuperscript{5}Tsinghua University}

\begin{document}
\maketitle
\begin{abstract}
Diffusion models have achieved impressive success in generating photorealistic images, but challenges remain in ensuring precise semantic alignment with input prompts. Optimizing the initial noisy latent offers a more efficient alternative to modifying model architectures or prompt engineering for improving semantic alignment. A latest approach, InitNo, refines the initial noisy latent by leveraging attention maps; however, these maps capture only limited information, and the effectiveness of InitNo is highly dependent on the initial starting point, as it tends to converge on a local optimum near this point. To this end, this paper proposes leveraging the language comprehension capabilities of large vision-language models (LVLMs) to guide the optimization of the initial noisy latent, and introduces the Noise Diffusion process, which updates the noisy latent to generate semantically faithful images while preserving distribution consistency. Furthermore, we provide a theoretical analysis of the condition under which the update improves semantic faithfulness. Experimental results demonstrate the effectiveness and adaptability of our framework, consistently enhancing semantic alignment across various diffusion models. The code is available at https://github.com/Bomingmiao/NoiseDiffusion.
\end{abstract}    
\section{Introduction}
Diffusion models have become the dominant approach in text-to-image generation due to their exceptional performance in this area \cite{saharia2022photorealistic, ramesh2022hierarchical, rombach2022high}. The core mechanism of diffusion models involves progressively adding noise to an image and using a neural network, such as U-Net \cite{ronneberger2015u} or DiT \cite{peebles2023scalable}, to predict the noise and then remove it based on the prediction during the reverse process. In the latest diffusion models, denoising is guided by conditional text embeddings from input prompts.  Recent advancements have also focused on performing the diffusion process in a lower-dimensional latent space. This latent space is typically derived from a pre-trained encoder, such as a variational autoencoder (VAE) \cite{kingma2013auto}, which enhances both efficiency and image consistency. 

Despite their success in generating high-quality images, diffusion models still face challenges in producing images that align with the description of the prompts. Improving diffusion models \cite{balaji2022ediff,ramesh2022hierarchical,saharia2022photorealistic,xue2024raphael} and implementing prompt engineering \cite{wang2023reprompt,liu2022design,brade2023promptify,pavlichenko2023best} represent two well-established approaches to addressing this challenge. While the former typically involves substantial training costs, the latter, though capable of enhancing generation quality through refined input prompts, may occasionally diverge semantically from the original prompts. In contrast, optimizing latent input remains relatively underexplored. \citet{chefer2023attend} observe that cross-attention maps can reveal the degree of alignment between text prompts and generated images, and propose a method to optimize latent variables at each denoising step. Building on an in-depth examination of attention layers within diffusion models, \citet{guo2024initno} introduce InitNo, an approach that optimizes the initial noisy latent variables and outperforms existing methods in generating semantically accurate images.

\begin{figure}[htbp]
  \centering
   \includegraphics[width=0.98\linewidth]{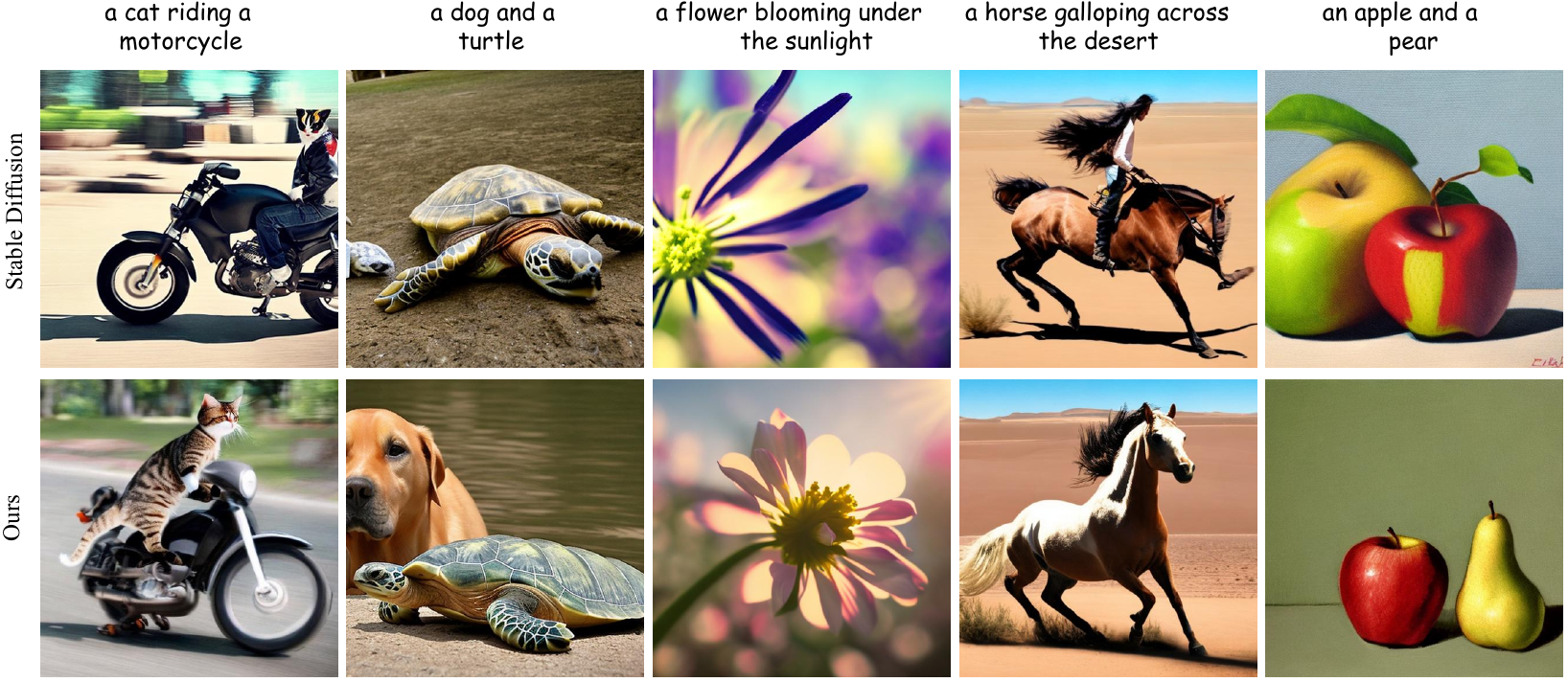}
   \caption{Example results of Stable Diffusion models and ours. Given a fixed initial noisy latent, we optimize the latent toward an area that can generate images aligned with the input prompts.}
   \label{fig:examples}
\end{figure}

However, the InitNo framework still faces two notable challenges. First, InitNo is effective only when the target point lies within the \(\ell_\infty\) neighborhood of the initial point. This is because InitNo uses a gradient-based method to update the mean and variance of the latent variable, which may lead to a distribution shift. To prevent excessive deviation of the updated latent variable distribution from the standard Gaussian distribution, InitNo restricts its search for the optimal solution to a local region near the initial point. Second, the misalignment between the generated images and input prompts often arises from insufficient understanding of the prompt content and image components \cite{liu2024towards}, suggesting that relying solely on attention layers to improve alignment may have limited effectiveness.

In response to these challenges, this paper presents a novel framework for optimizing the initial noisy latent during the denoising process. To improve the generation process's understanding of image components and prompt content, we employ a large vision-language model (LVLM) for supervision. Specifically, the image generated by the diffusion model, along with a question derived from the prompt, is input into the LVLM to compute the Visual Question Answering (VQA) score proposed by \cite{lin2025evaluating}. We then optimize the latent variable to maximize this score to ensure better alignment between the generated image and the prompt. Regarding the preservation of distribution consistency during latent variable optimization, we introduce the Noise Diffusion method. This approach applies the diffusion forward process to update the latent variable, progressively adding Gaussian noise to the original latent variable, shifting it to a region conducive to generating more semantically consistent and satisfactory images. The challenge with this method lies in determining the optimal noise that can increase the VQA score, as existing methods are ineffective in using gradient information to sample Gaussian noise. To address this, we sample a set of noises at each iteration, compute the step difference based on the pre-defined step size and the sampled noise, and utilize gradient information to select the most appropriate step difference for updating the latent variable. We theoretically prove that when the ratio of the inner product of the step difference and gradient to the square of the step difference's norm exceeds a threshold, the optimization will increase the VQA score. As shown in Fig.~\ref{fig:examples}, when the vanilla Stable Diffusion (SD) model fails to produce images that align with the prompt, our method for refining the latent variable ensures better alignment between the generated images and the prompts. Our contributions are summarized as follows:
\begin{itemize}
    \item We propose a novel framework that harnesses the semantic understanding of LVLMs to supervise the diffusion generation process and introduces the Noise Diffusion method to optimize the initial noisy latent variables while preserving the distribution of the latent variables, thereby enhancing the semantic faithfulness of the generated images to the input prompts.
    \item We provide a theoretical analysis of the condition under which updating the latent variables can increase the VQA score. Based on this analysis, we propose a strategy for selecting the noise for update process using gradient information.
    \item Extensive experimental results demonstrate the superiority of our method, which can seamlessly and effectively enhance the semantic faithfulness of various diffusion models.
\end{itemize}
\section{Related Work}
Early studies on text-to-image synthesis mainly focus on GANs \cite{zhang2017stackgan,xu2018attngan,zhu2019dm, zhang2021cross, tao2022df} and auto-regressive models \cite{ding2021cogview, ramesh2021zero, yu2022scaling, chang2023muse}. More recently, diffusion models have outperformed these methods \cite{dhariwal2021diffusion}. The concept of training a generative model by adding noise to data and learning the reverse process to restore the original data distribution is first proposed by \cite{sohl2015deep}. DDPM \cite{ho2020denoising} showcases the remarkable ability of diffusion models for unconditional image generation. \citet{song2020score} extend this by presenting a stochastic differential equation (SDE) to model the forward and reverse processes, and derive an equivalent neural ordinary differential equation (ODE) that samples from the same distribution, enabling exact likelihood computation and improving sampling efficiency. To achieve photorealism in class-conditional settings, \citet{dhariwal2021diffusion} enhance diffusion models with classifier guidance by training a classifier on noisy images and using its gradients to guide samples toward the target label. \citet{ho2022classifier} introduce classifier-free guidance, which combines the score estimates from a jointly trained conditional and unconditional model. For generating images from free-form textual prompts, \citet{nichol2021glide} employ a text encoder to condition the denoising process on language descriptions, and demonstrate that text-guided diffusion models with classifier-free guidance can yield higher quality images. 

Meanwhile, numerous efforts have been made to address the issue of unfaithfulness in image generation using diffusion models. One research direction is to explore how to train more powerful diffusion models. \citet{balaji2022ediff} observe the difference of the importance of the text condition in different denoising stages and propose training an ensemble of text-to-image diffusion models specialized for different synthesis stages. \citet{ramesh2022hierarchical} enhance the output of diffusion models with prior CLIP image embeddings added to the timestep embedding of diffusion models and concatenated with the output of text encoder. \citet{saharia2022photorealistic} enhance the alignment of textual and visual content with the language understanding of large language models (LLMs). \citet{xue2024raphael} introduce stacking tens of mixture-of-experts (MoEs) layers to generate highly artistic images. \citet{segalis2023picture} point out relabeling the corpus with a specialized automatic captioning model can significantly improve the quality of the dataset for training a diffusion model. 

Another line of works explore strategies without retraining or modifying the models. \citet{liu2022compositional} propose generating an image with a set of diffusion models, each of which model a certain component of the image. \citet{si2024freeu} observe differences in feature contributions between the U-Net architecture's main backbone and its skip connections, and improve the generation quality by introducing two specialized modulation factors for the feature maps in these respective components. \citet{feng2022training} propose spliting the input text prompts and manipulating the cross-attention representations to better preserve the compositional semantics. \citet{chefer2023attend} introduce the concept of Generative Semantic Nursing (GSN), intervening in the denoising process by optimizing the latent at each timestep to refine the cross-attention, so as to encourage generating all subjects described in the text prompt. \citet{agarwal2023star, li2023divide} further improve the optimization objective for better update direction. \citet{guo2024initno} point out that optimizing the noisy latent at each denoising timestep requires carefully designed parameters and is hard to control the extent of optimization, and propose an alternative method by adjusting the sampled noise in the initial latent space.

\begin{figure}[ht]
  \centering
   \includegraphics[width=1.0\linewidth]{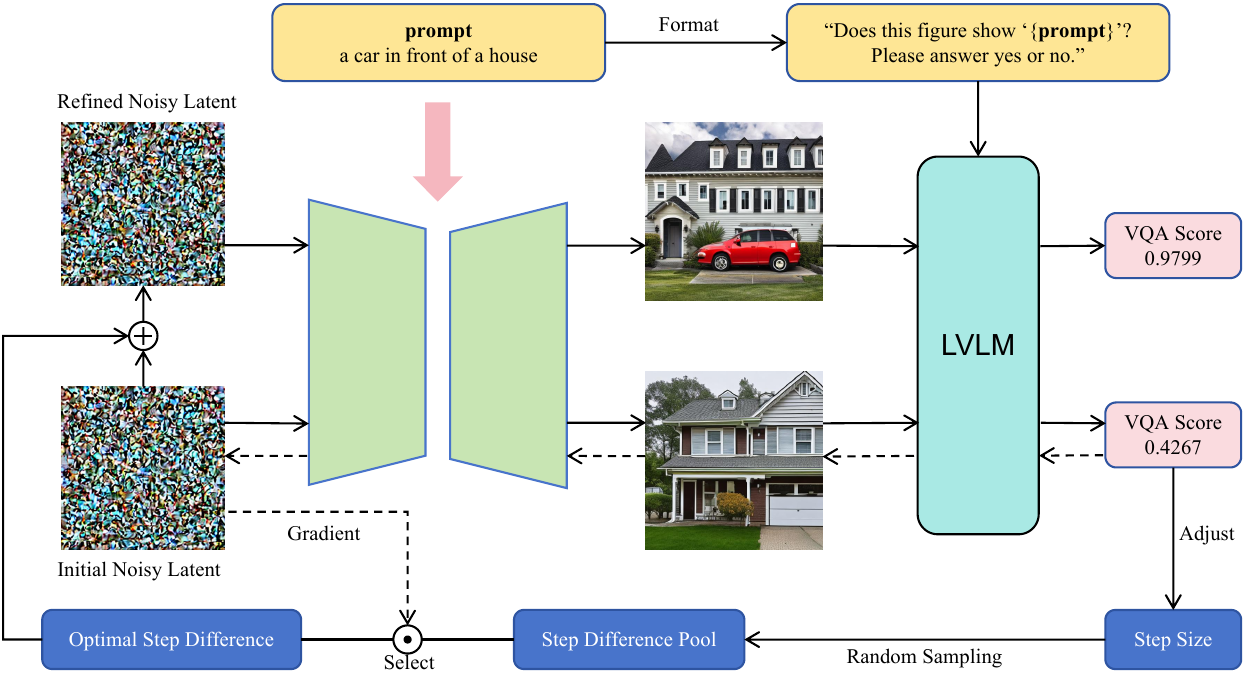}
   \caption{The framework of Noise Diffusion. The image generated from the initial noisy latent is fed into the LVLM along with a question formatted as, ``Does this figure show `\{\textbf{prompt}\}'? Please answer yes or no." The probability of the token ``Yes" serves as the VQA score. The step size for updating the noisy latent is dynamically adjusted based on the score value. Gradient information is then used to select the optimal noise for the update according to the step difference.
}
   \label{fig:framework}
\end{figure}
\begin{figure*}[t]
  \centering
  \includegraphics[width=0.9\linewidth]{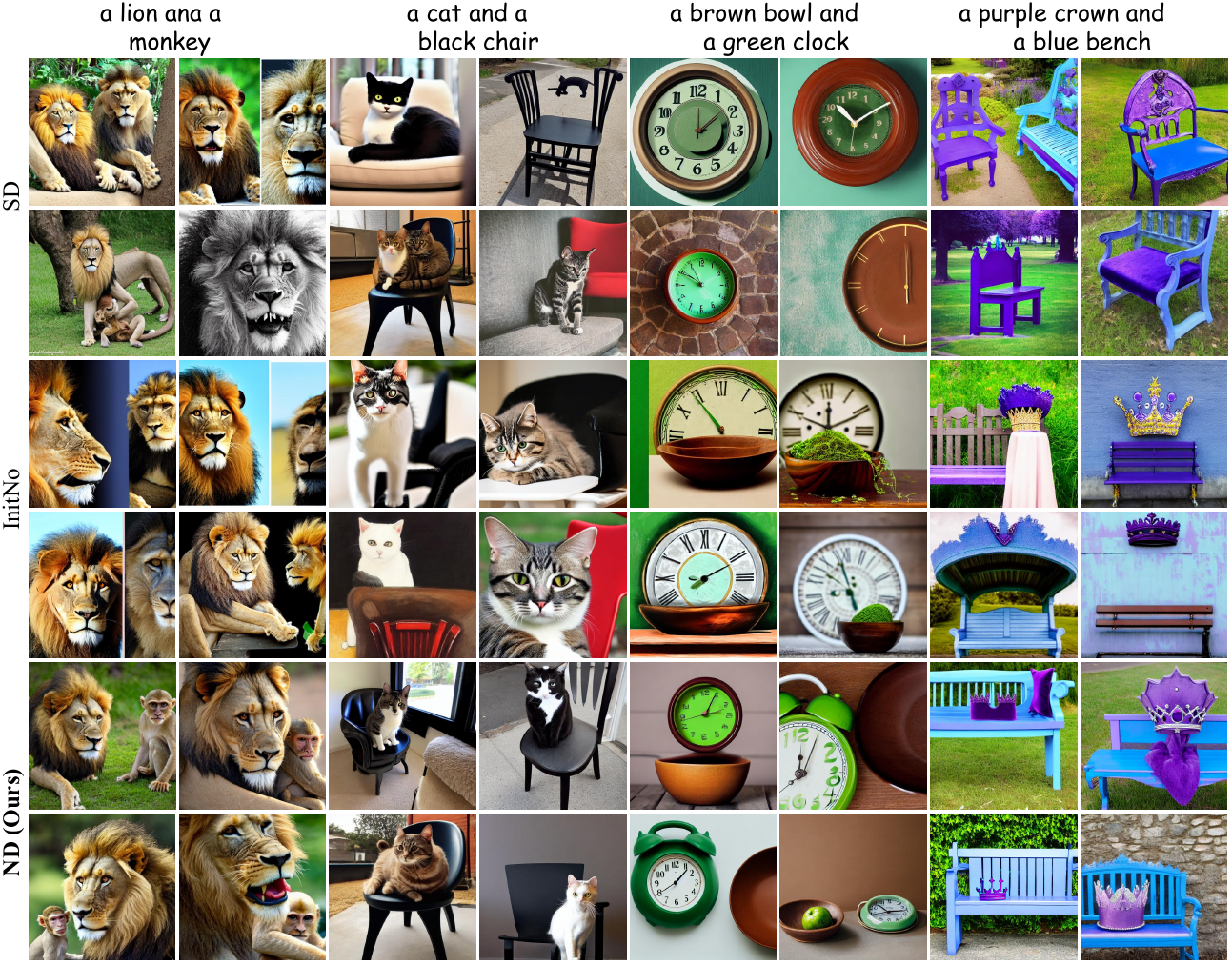}
   \caption{Qualitative comparison for simple cases. Each image is generated with the same prompt and random seed for all methods. The images generated by our method contain objects that most closely match the features described in the prompts.}
   \label{fig:simple}
\end{figure*}
\section{Method}
\subsection{Preliminary}
\textbf{VQA Score.} We utilize VQA score proposed by \cite{lin2025evaluating} to measure the alignment between generated images and input prompts. The image $I$ is input into a large vision-language model along with the question: “Does this figure show `\{\textbf{prompt}\}'? Please answer yes or no.” The VQA score is then computed as:
\begin{equation}
    VQA(I,\textbf{prompt})=P(``Yes" \mid I,\textbf{prompt}).
\end{equation}

\noindent
\textbf{Diffusion Models.} For diffusion process in latent space, the noise is added progressively throgh a variance schedule $\beta_1,\ldots,\beta_T$. Let $\alpha_t:=1-\beta_t$, $\bar{\alpha}_t:=\prod_{s=1}^t \alpha_s$, the noisy latent can be expressed as:
\begin{equation}
q\left(z_t \mid z_0\right)=\mathcal{N}\left(z_t ; \sqrt{\bar{\alpha}_t} z_0,\left(1-\bar{\alpha}_t\right) \mathbf{I}\right)
\end{equation}
For text-to-image synthesis, the denoising process begins from the initial latent variable $z_T$. Let $\mathcal{C}$ denote the text embedding and $\varnothing$ denote the null embedding, the classifier-free noise prediction can be expressed as:
\begin{equation}
\epsilon_\theta\left(z_t, t, \mathcal{C}, \varnothing\right)=w \cdot \epsilon_\theta\left(z_t, t, \mathcal{C}\right)+(1-w) \cdot \epsilon_\theta\left(z_t, t, \varnothing\right),
\end{equation}
where $w$ denotes the guidance scale parameter, which is fixed as 7.5 in Stable Diffusion models. For simplicity, we use $\epsilon_\theta(z_t)$ to represent $\epsilon_\theta(z_t,t,\mathcal{C}, \varnothing)$. Since noisy latent optimization requires a deterministic denoising process, we adopt DDIM \cite{song2020denoising} to ensure consistency across the iterations. In the DDIM framework, the noisy latent from the previous timestep can be obtained by:
\begin{equation}
\begin{aligned}
    z_{t-1}&=\sqrt{\frac{\alpha_{t-1}}{\alpha_t}}\left(z_t-\sqrt{1-\alpha_t} \epsilon_\theta\left(z_t\right)\right) \\
    &+\sqrt{1-\alpha_{t-1}} \epsilon_\theta\left(z_t\right).
\end{aligned}
\end{equation}
After completing the denoising process, a decoder $\mathcal{D}(\cdot)$ is applied to the final denoised latent $z_0$ to generate the output image $I$: $I = \mathcal{D}(z_0)$.

\noindent
\textbf{Gradient Approximation.} In our pipeline, $z_T$ is denoised to $z_0$, and $z_0$ is decoded into image $I$, which is fed to the LVLM to compute the VQA score. Therefore the score of $z_T$ can be defined as:
\begin{equation}
    s(z_T) =  VQA(\mathcal{D}(\Omega(z_T,\mathcal{C},\varnothing)), \textbf{prompt}),
    \label{eq:score}
\end{equation}
where $\Omega$ denotes the DDIM denoising process. The gradient of the objective with respect to $z_T$ can be computated as:
\begin{equation}
    \nabla_{z_T}s(z_T) = \frac{\partial s(z_T)}{\partial I}\frac{\partial I}{\partial z_0} \frac{\partial z_0}{\partial z_T},
\end{equation}
where $\frac{\partial z_0}{\partial z_T}$ constitutes the primary computational cost, as it requires back propagation through multiple timesteps in the denoising process. The derivative of the latent variable at the previous timestep with respect to the current timestep is given by:
\begin{equation}
    \frac{\partial z_{t-1}}{\partial z_t}=\sqrt{\frac{\alpha_{t-1}}{\alpha_t}}+\sqrt{\frac{\alpha_t-\alpha_{t-1}}{\alpha_t}} \frac{\partial \epsilon_\theta\left(z_t\right)}{\partial z_t}.
\end{equation}
According to the chain rule, $\frac{\partial z_0}{\partial z_T}$ can be expanded as:
\begin{equation}
    \begin{aligned}
    & \frac{\partial z_0}{\partial z_T} = \left( \sqrt{\frac{\alpha_0}{\alpha_1}} + \sqrt{\frac{\alpha_1 - \alpha_0}{\alpha_1}} \frac{\partial \epsilon_\theta\left(z_1\right)}{\partial z_1} \right) \frac{\partial z_1}{\partial z_2} \cdots \frac{\partial z_{T-1}}{\partial z_T} \\
    & = \left( \sqrt{\frac{\alpha_0}{\alpha_1}} \frac{\partial z_1}{\partial z_2} + \sqrt{\frac{\alpha_1 - \alpha_0}{\alpha_1}} \frac{\partial \epsilon_\theta\left(z_1\right)}{\partial z_2} \right) \frac{\partial z_2}{\partial z_3} \cdots \frac{\partial z_{T-1}}{\partial z_T} \\
    &  = \sqrt{\frac{1}{\alpha_T}} + \sum_{i=1}^{T} \sqrt{\frac{\alpha_i - \alpha_{i-1}}{\alpha_{i-1} \alpha_i}} \frac{\partial \epsilon_\theta\left(z_i\right)}{\partial z_T},
    \end{aligned}
\end{equation}
where $\alpha_0=1.$ The computational cost of the full process is extremely high. To simplify the calculation, we adopt the simple approximation proposed by \cite{miao2024advlogo}, where $\epsilon_\theta(z_t)$ is treated as a constant $\epsilon_t$ for all $t=1,\ldots,T.$ In this way $\frac{\partial \epsilon_\theta(z_t)}{\partial z_T}=0$ since $\epsilon_\theta(z_t)$ is assumed to be independent to $z_T$. Consequently, we have $\frac{\partial z_0}{z_T}=\sqrt{\frac{1}{\alpha_T}}.$ 
Therefore the gradient calculation can be simplified as:
\begin{equation}
    \nabla_{z_T}s(z_T) =\sqrt{\frac{1}{\alpha_T}} \frac{\partial s(z_T)}{\partial I}\frac{\partial I}{\partial z_0}.
\label{eq:gradient}
\end{equation}

\subsection{Noise Diffusion}
We use the same approach as the diffusion forward process to transfer the initial noisy latent to a new state. Specifically, for the current latent $z_T$, the updated latent $z'_T$ is obtained by: 
\begin{equation}
    z'_T = \sqrt{1 - \gamma} z_T + \sqrt{\gamma} \sigma, 
\label{eq:update}
\end{equation}
where $\sigma \sim \mathcal{N}(0, \mathbf{I})$, $\gamma$ is the step size. Since $z_T$ and $\sigma$ both cohere to standard Gaussian distribution, the updated latent $z'_T \sim \mathcal{N}(0,\mathbf{I})$ holds true, regardless of the value of $\gamma$. This allows us to obtain a latent that is distant from the original while remaining in the same distribution.

\noindent
\textbf{Score-Aware Step Size}. The step size determines the magnitude of the perturbation to the latent variable. Considering that the step size should be large when the score is low, and small when the score is high, we employ the following function to dynamically adjust the step size:
\begin{equation}
    \gamma=1-\sqrt{s(z_T)}.
\end{equation}

\begin{algorithm}
\caption{Noise Diffusion}
\label{alg:noisediffusion}
\begin{algorithmic}[1]
\REQUIRE A prompt $P$, a pretrained diffusion model and LVLM, number of denoising steps $T$, initial latent at the last timestep $z_T$, max optimization epoch $M$, number of candidate noises $N$.
\ENSURE The refined image
\STATE $\mathcal{C},\varnothing \gets \textbf{TextEncoder}(P,``").$
\STATE $I \gets \mathcal{D}(\Omega(z_T,\mathcal{C},\varnothing))$.
\STATE Calculate $s(z_T)$ based on Eq. (\ref{eq:score}).
\STATE $I^*\gets I, s^*\gets s(z_T).$
\FOR{$m=1, \ldots, M$}
    \STATE $\gamma = 1-\sqrt{s(z_T)}$.
    \STATE Calculate $\nabla_{z_T}s(z_T)$ based on Eq. (\ref{eq:gradient}).
    \STATE Randomly sample a set of noises $[\sigma_1,\ldots,\sigma_N]$.
    \STATE $v_i = (\sqrt{1-\gamma}-1)z_T + \sqrt{\gamma}\sigma_i, i=1,\ldots,N.$
    \STATE $\sigma = \underset{i \in [1,\ldots,N]}{\arg \max}\nabla_{z_T}s(z_T)v_i/\|v_i\|_2^2$.
    \STATE $z_T \gets \sqrt{1 - \gamma} z_T + \sqrt{\gamma} \sigma.$
    \STATE $I \gets \mathcal{D}(\Omega(z_T,\mathcal{C},\varnothing)).$
    \IF{$s(z_T)>s^*$}
    \STATE $s^*\gets s(z_T), I^* \gets I.$
    \ENDIF
\ENDFOR
\STATE \textbf{return} $I^*$
\end{algorithmic}
\end{algorithm}

\noindent
\textbf{Noise Selection.} Let $v$ denote the step difference between $z_T$ and $z'_T$ obtained by Eq. (\ref{eq:update}). $v$ can be expressed as:
\begin{equation}
    v = (\sqrt{1-\gamma}-1)z_T + \sqrt{\gamma} \sigma.
\end{equation}
For simplicity, we regard $z_T$ as a vector, the score of updated latent variable $s(z'_T)$ can be expressed in Taylor expansion:
\begin{equation}
\begin{aligned}
    s(z'_T) &= s(z_T) + \nabla_{z_T}s(z_T) v
    + \frac{1}{2} v^T \nabla_\xi^2 s(\xi)v,
\end{aligned}
\end{equation}
where $\xi = z_T + \rho v, \rho \in (0,1)$. Therefore, $\nabla_{z_T}s(z_T)v$ plays an important role in determining the value of $s(z'_T)$. Randomly sampling noise \(\sigma\) from a Gaussian distribution may not guarantee that \(\nabla_{z_T}s(z_T)v\) is positive, and thus this may not guarantee an overall increase in the score. To ensure that the overall update process progresses toward regions with higher VQA scores, we use the gradient information to select the appropriate noise for updating. As presented in Alg.~\ref{alg:noisediffusion}, once the step size is obtained based on the score and the gradient is computed, we then sample a set of candidate noises, and select the noise that maximizes $\nabla_{z_T}s(z_T)v/\|v\|_2^2$. Finally, we update $z_T$ using the selected noise by Eq. (\ref{eq:update}). We provide an overview of our proposed Noise Diffusion method in Fig. \ref{fig:framework}.

\noindent
\textbf{Feasibility Analysis.} To obtain a latent that yields a higher score after iterations, the noise for updating $z_T$ must satisfy the certain condition. Existing widely adopted activation functions in neural networks have bounded second derivatives almost everywhere \cite{dubey2022activation}. So there exists a positive constant $c$ so that for any given $\xi$, 
\begin{equation}
    \|\nabla_\xi^2 s(\xi)\|_2 \overset{\text{a.e.}}{\leq} c
\end{equation}
 is satisfied. Then if the condition $\nabla_{z_T}s(z_T) v / \|v\|_2^2 \geq \frac{c}{2}+\delta, \delta>0$ holds true, we can obtain that
\begin{equation}
\begin{aligned}
     s(z'_T) &= s(z_T) + \nabla_{z_T}s(z_T) v + \frac{1}{2} v^T \nabla_{z_T}^2 s(\xi) v \\
     & \overset{\text{a.e.}}{\geq} s(z_T)+ \nabla_{z_T}s(z_T) v - \frac{c}{2}\|v\|_2^2 \\
     & \geq s(z_T) + \frac{c + \delta}{2}\|v\|_2^2 - \frac{c}{2}\|v\|_2^2 \\
     & = s(z_T) + \delta \|v\|_2^2.
\end{aligned}
\end{equation}
Therefore, maximizing $\nabla_{z_T}s(z_T)v/\|v\|_2^2$ is effective in identifying the noise that satisfies the condition for updating towards a higher score.

\section{Experiment}
\begin{figure*}[t]
  \centering
  \includegraphics[width=0.9\linewidth]{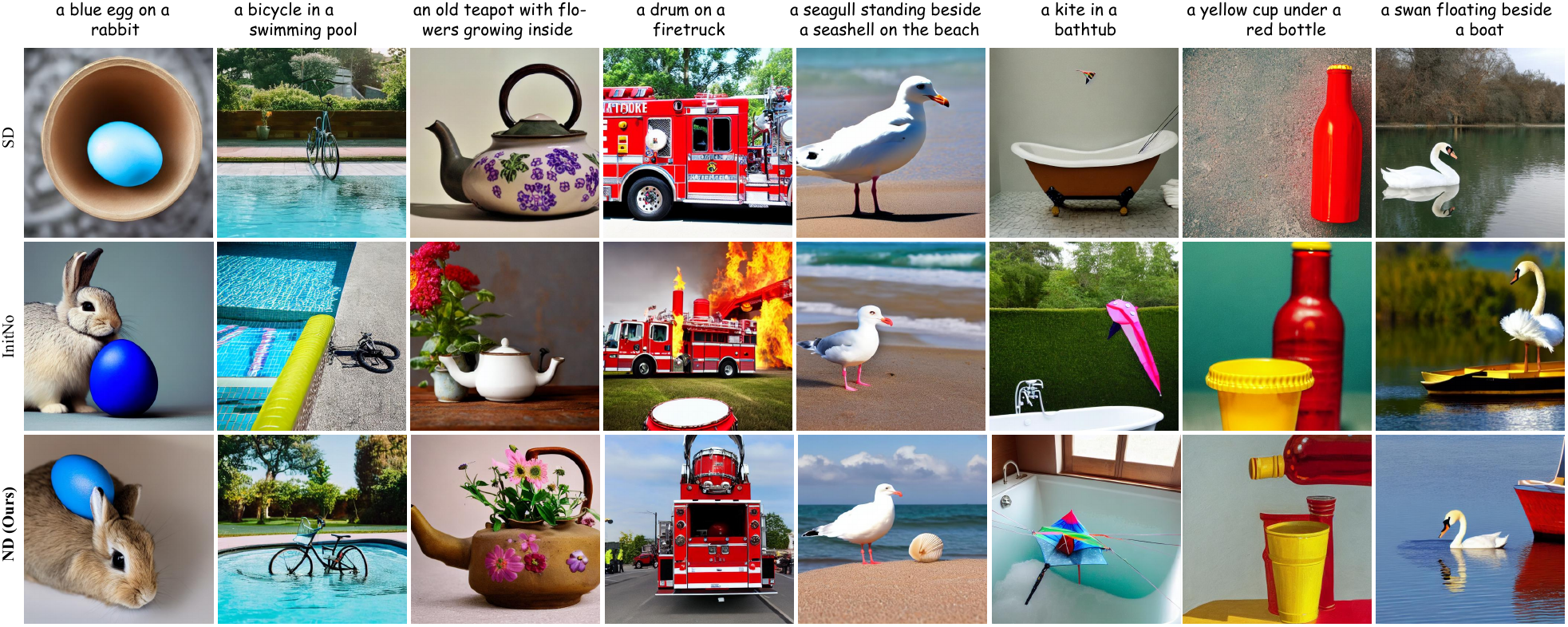}
   \caption{Qualitative comparison for complex cases. Each image is generated with the same text prompt and random seed for all methods. Our method exhibits strong understanding of the positional relationships described in the prompts.}
   \label{fig:complex}
\end{figure*}
\begin{figure}[ht]
  \centering
   \includegraphics[width=0.95\linewidth]{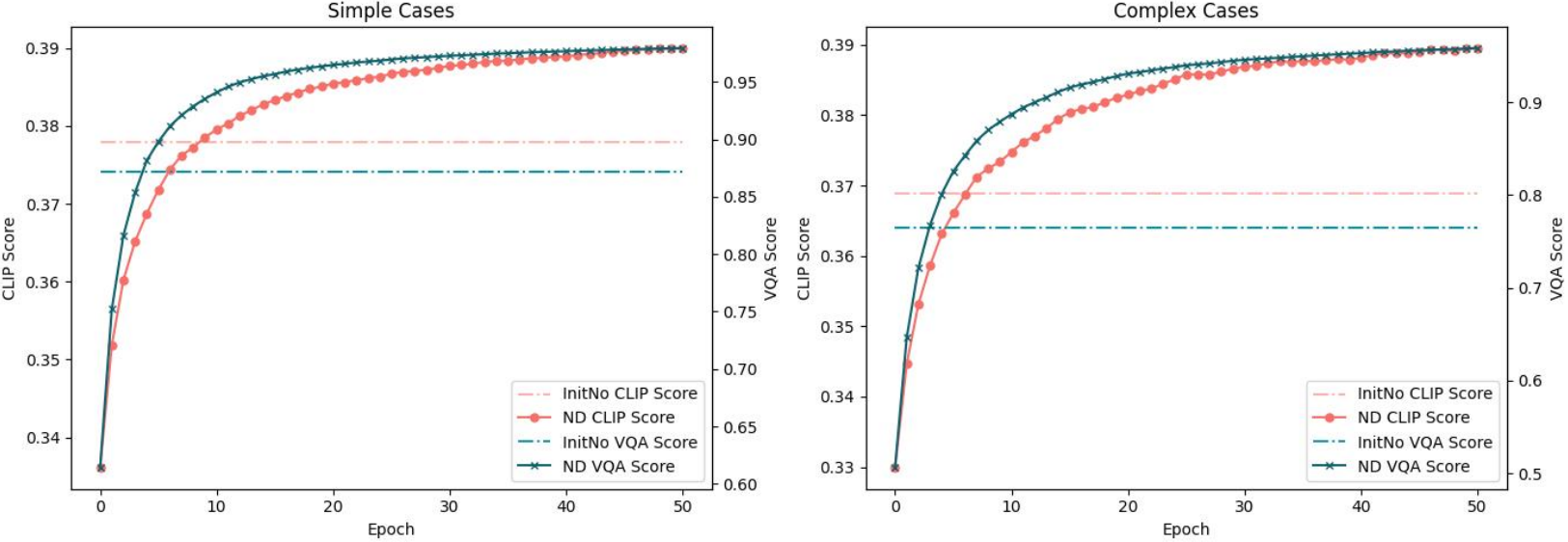}
   \caption{The average CLIP score and VQA score of the generated images both increase as the epochs progress and eventually converge. Compared to InitNo, the Noise Diffusion (ND) method consistently outperforms InitNo in both CLIP and VQA scores.}
   \label{fig:quantative_results}
\end{figure}
\begin{figure*}
  \centering
  \includegraphics[width=0.9\linewidth]{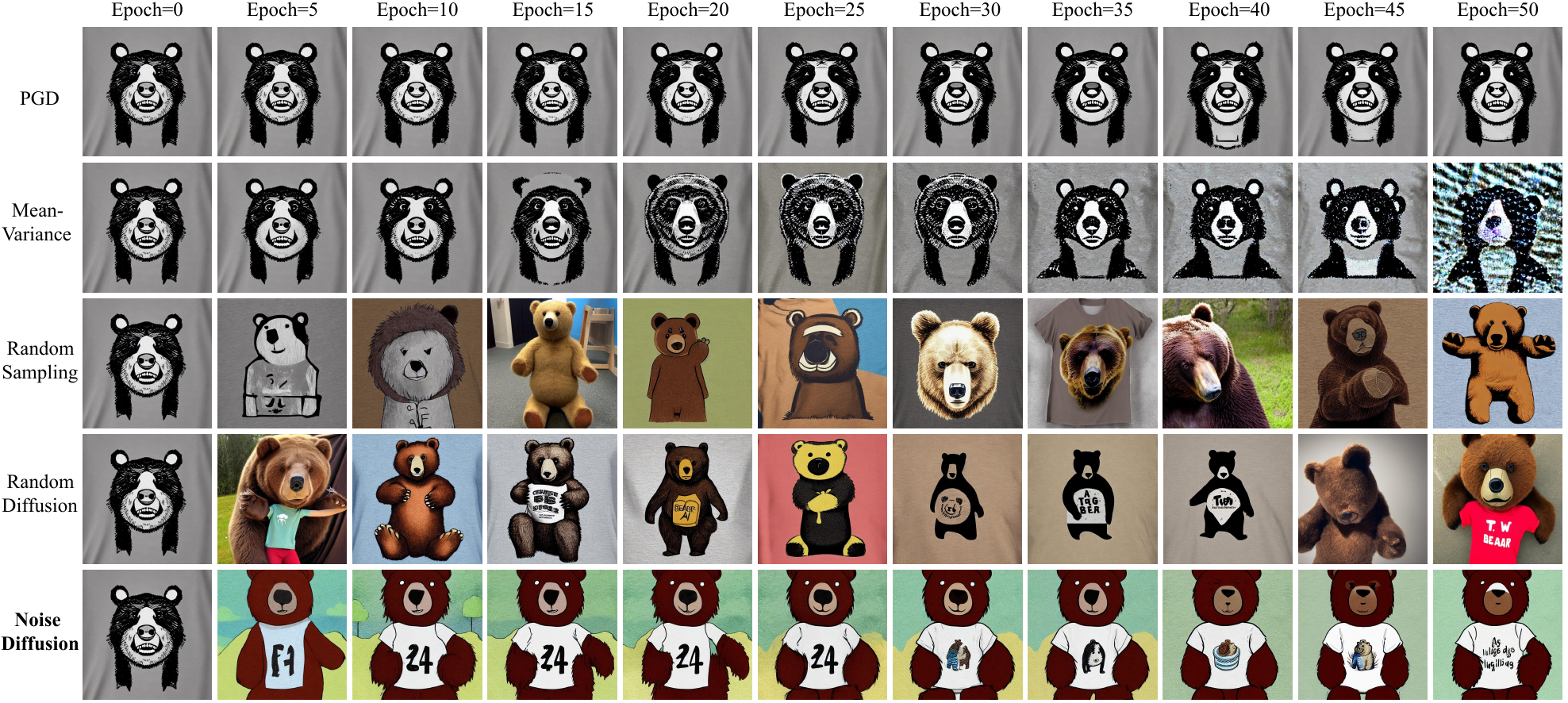}
   \caption{We compare different optimization methods using the prompt ``a bear wearing a T-shirt" as an example. PGD and Mean-Variance can only converge to a local optimum. Random Sampling searches for the optimal point randomly and proves ineffective. Random Diffusion improves upon Random Sampling by introducing an adjusted step size, making it more effective. However, Noise Diffusion outperforms all other methods, efficiently producing an image that closely matches the prompt.}
   \label{fig:comparison}
\end{figure*}

\begin{figure}[ht]
 \centering
  \includegraphics[width=0.9\linewidth]{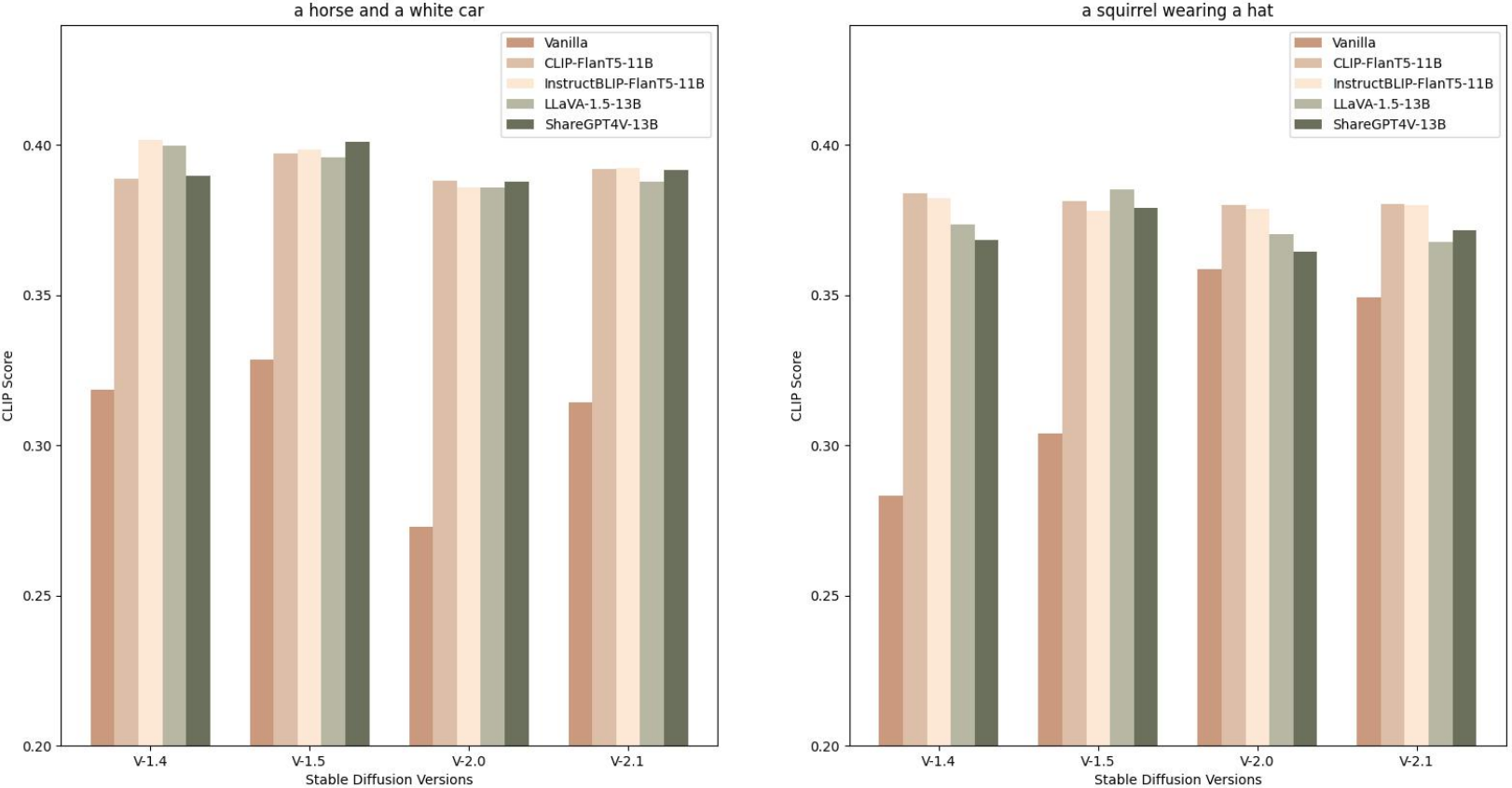}
   \caption{Comparison of the average CLIP score of images generated by different SD models and refined by various LVLMs.}
   \label{fig:quant_ablation}
\end{figure}
\begin{figure*}[ht]
  \centering
  \includegraphics[width=0.9\linewidth]{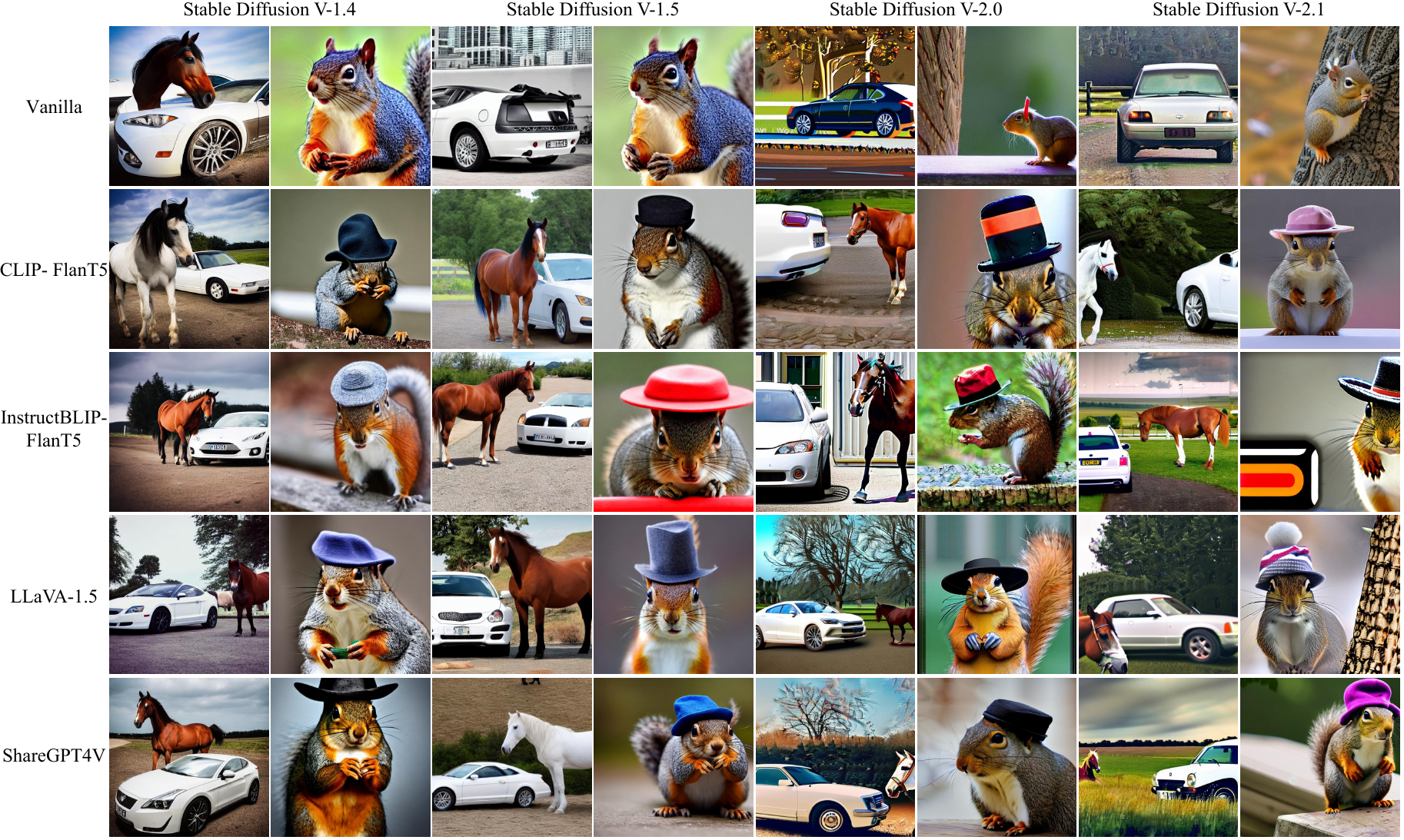}
   \caption{We use the prompts ``a horse and a white car" and ``a squirrel wearing a hat" as examples, conducting experiments on Stable Diffusion V-1.4, V-1.5, V-2.0, and V-2.1. We apply CLIP-FlanT5-11B, InstructBLIP-FlanT5-11B, LLaVA-1.5-13B, and ShareGPT-4V-13B as the LVLMs for refining the generation. The faithfulness of image generation improves across all combinations of models.}
   \label{fig:ablation}
\end{figure*}
\subsection{Implementation Details}
\textbf{Model Choice.} In accordance with \cite{guo2024initno}, we utilize the official Stable Diffusion V-1.4 \cite{rombach2022high} as the base diffusion model alongside CLIP-FlanT5-11B  \cite{lin2025evaluating} to compute the VQA score in Sec.~\ref{subsec:qualitative}, Sec.~\ref{subsec:quantitative} and Sec.~\ref{subsec:optimization}. In Sec.~\ref{subsec:models}, we evaluate the effectiveness and generalization capability of our method across various diffusion models, including Stable Diffusion V-1.5, V-2.0, and V-2.1, as well as several LVLMs, such as InstructBLIP-FlanT5-11B \cite{dai2023instructblipgeneralpurposevisionlanguagemodels}, LLaVA-1.5-13B \cite{liu2024visual}, and ShareGPT4V-13B \cite{chen2023sharegpt4v}. We compute the CLIP score using the CLIP ViT-L/14 model \cite{radford2021learning}. 

\noindent
\textbf{Hyper-parameters.} The number of denoising steps $T$, max optimization epoch $M$ and candidate noises $N$ are all set as 50. Evaluated on a single NVIDIA RTX A6000 (48GB), SD V-1.4 synthesizes an image in an average of 6.08 seconds, while our method requires additional 6.71 seconds for one optimization epoch. Although our method requires additional computational cost, it significantly improves the semantic faithfulness of the generated images. Moreover, choosing smaller hyper-parameters can also reduce the computational cost. We discuss the impact of hyper-parameters in the Appendix. B.

\noindent
\textbf{Datasets.} We use two prompt datasets, covering simple and complex cases to evaluate the alignment of prompts and images. The simple cases focus on object structures and are sourced from the dataset proposed by \cite{chefer2023attend}. For complex cases, we retain the objects from the simple cases, randomly replacing conjunctions such as ``and" or ``with" with spatial terms like ``on," ``under," ``above," or ``below" to represent positional relationships. See the Appendix. C for detailed descriptions of the datasets.

\subsection{Qualitative Comparison}
\label{subsec:qualitative}
Fig.~\ref{fig:simple} and Fig.~\ref{fig:complex} illustrate a comparison between our proposed method, Stable Diffusion (SD) and InitNo under simple and complex cases, respectively. With identical prompts and seeds, our method demonstrates clear advantages over the other approaches. In Fig.~\ref{fig:simple}, we observe that images generated by the vanilla Stable Diffusion model may exhibit issues such as object blending, object omission, or color distortion. In mitigating these deficiencies, our method demonstrates clear superiority over the existing state-of-the-art approach, InitNo. Specifically, in the case of the prompt ``a lion and a monkey," where SD fails to render the ``monkey," InitNo also struggles to capture the features of the monkey. In contrast, our method encourages the diffusion model to generate a harmonious image containing both the lion and the monkey. Furthermore, with the prompt ``a brown bowl and a green clock," while InitNo successfully outlines the structures of the bowl and the clock, it frequently misinterprets their colors. Our method, however, exhibits more effective performance in accurately rendering the colors.

When we introduce more complex cases, we find that although InitNo may succeed in rendering the structure of the objects described in the prompts, it fails to capture the underlying logic, leading to a significant deviation from the required semantics. For example, with the prompt ``a blue egg on a rabbit," SD fails to depict a rabbit. While InitNo successfully presents both a rabbit and a blue egg, it completely overlooks the concept of ``on," resulting in an image of ``a blue egg beside a rabbit." In contrast, our method effectively adjusts the relationship between the rabbit and the egg, demonstrating the advantage of incorporating feedback from the LVLM to refine image generation.

\subsection{Quantitative Comparison}
\label{subsec:quantitative}
To evaluate the effectiveness of our proposed method, we conduct quantitative comparative experiments on two datasets. For each prompt, we randomly select 25 seeds. To reflect the changes in semantic faithfulness during the optimization process of our method, we plot the average VQA and CLIP scores of the generated images at different epochs across two datasets in Fig.~\ref{fig:quantative_results}. Additionally, we also present the average VQA and CLIP scores after optimization using the InitNo method for comparison. Both the VQA score and the CLIP score of the generated images show an ascending trend over time. At epoch 0, the images are generated using the unoptimized Vanilla SD model. On datasets with simple cases, when the optimization epoch reaches 10, our method achieves a VQA score of 0.9352, surpassing the average VQA score of InitNo (0.8717). As the optimization epoch approaches 50, the performance improvements brought by our method begin to plateau, with the VQA score reaching 0.9790, exceeding the average VQA score of InitNo by 0.1073. On datasets with complex cases, the performance advantage of our method is even more pronounced. At epoch 10, our method achieves a VQA score of 0.8791, surpassing the average VQA score of InitNo by 0.1144. As the optimization epoch approaches 50, our method achieves a VQA score of 0.9578, again surpassing the average VQA score of InitNo by 0.1930. This underscores the clear advantages of our method in handling complex semantic information.

\subsection{Comparison of Different Optimization Techniques}
\label{subsec:optimization}
Our method consists of two key components. The first is the adoption of a large vision-language model to refine the generation, which as discussed earlier, demonstrates the superior ability to capture the detail of images. The second part is the proposed Noise Diffusion method. To highlight the advantages of our proposed optimization technique, we examine a specific example using the prompt ``a bear wearing a T-shirt." As shown in Fig.~\ref{fig:comparison}, we present the images generated by the optimized noisy latent at different epochs. Among these comparative methods, Mean-Variance refers to the perturbation method used in InitNo, where the mean and variance of the Gaussian distribution are adjusted using Adam optimizer \cite{kingma2014adam}. For a comprehensive comparison, we also introduce the PGD method \cite{madry2017towards}, which directly perturbs the latent, as well as Random Sampling, where we randomly select another noise to replace the current noise at each timestep. Additionally, we introduce Random Diffusion, where the step size is determined by the VQA score, but the noise is randomly sampled for optimization rather than being guided by the gradient as in our proposed Noise Diffusion. As illustrated in Fig.~\ref{fig:comparison}, the content in the initial image deviates significantly from the prompt description. Methods like PGD, which directly perturbs the latent, or Mean-Variance, which perturbs the mean and variance based on gradients, show limited changes to the image structure, with relatively small visual adjustments. By the 50th epoch, the visual quality of images produced by the Mean-Variance method has notably declined. This suggests that the initial latent lies in a low-score region within the noise space, and there are no nearby points with significantly higher scores. Consequently, this optimization faces limitations due to the trade-off between visual quality and perturbation extent, making it challenging to achieve substantial improvement. In contrast, methods such as Random Sampling, Random Diffusion, and our proposed Noise Diffusion demonstrate global optimization. Compared to the initial image, the appearance of the bear evolves considerably over subsequent epochs. Random Sampling, which operates with a fixed step size of 1 at each iteration, depends on the proportion of the latent space that yields favorable images. If this proportion is low, Random Sampling becomes highly inefficient. Random Diffusion enhances this approach by introducing a dynamic step size adjustment based on the VQA score. Although still not fully matching the prompt, the images generated by Random Diffusion align better with the prompt compared to those produced by Random Sampling. This validates the advantage of our step size adjustment method based on the score. Comparing Random Diffusion with Noise Diffusion, we observe that Noise Diffusion produces an image closely matching the prompt as early as epoch 5, and this result remains stable. This demonstrates the efficiency gained by incorporating gradient information into the update process.

\subsection{Impact of Models}
\label{subsec:models}
To evaluate the generalization capability of our method, we conduct experiments to assess the impact of different model configurations on the results. Specifically, we use CLIP-FlanT5-11B, InstructBlip-FlanT5-13B, LLaVA-1.5-13B and ShareGPT4V-13B to refine the text-to-image synthesis outputs of SD models V-1.4, V-1.5, V-2.0, and V-2.1. For evaluation, we select two prompts—``a horse and a white car" as a simple case and ``a squirrel wearing a hat" as a complex case—and generate images using 25 randomly selected seeds. As shown in Fig.~\ref{fig:quant_ablation}, the average CLIP scores of the generated images refined by LVLMs exceeds those of the vanilla settings, demonstrating that our method improves performance across all the models used, validating its effectiveness and generalization capability. Fig.~\ref{fig:ablation} highlights that SD fails to generate the concept of ``horse" in the prompt ``a horse and a white car" and ``hat" in the prompt "a squirrel wearing a hat." In contrast, the refined images successfully present the missing objects, further emphasizing the effectiveness of our optimization approach.

\section{Conclusion}
This paper proposes a novel framework that leverages the capabilities of large vision-language models (LVLMs) in image components and textual semantic understanding to optimize the initial noisy latent variables, thereby enhancing the semantic faithfulness of text-to-image synthesis. Specifically, we employ the Visual Question Answering (VQA) score to assess the alignment between the generated image and the textual prompt. Additionally, we introduce a noise diffusion approach, which updates the latent variables through a forward diffusion process, dynamically adjusting the step size and selecting the optimal noise for latent variable update based on the approximate gradient of the VQA score. We theoretically analyze the condition under which the update can enhance the VQA score. Our approach surpasses the current state-of-the-art in generating semantically faithful images, offering a flexible, plug-and-play solution compatible with existing diffusion models for training-free controllable generation.

{
    \small
    \bibliographystyle{ieeenat_fullname}
    \bibliography{main}
}
\clearpage
\appendix
\setcounter{section}{0}
\section{Theoretical Analysis}
In this section, we present the proof of the inequality $\|\nabla_{\xi}^2s(\xi)\|_2\overset{a.e.}{\leq}c.$ First, we prove that this inequality holds within the domain of the function $s(\cdot)$. Then, we prove that the region where the derivatives of the function are undefined forms a set of zero measure, which guarantees the effectiveness of our method.

\begin{table*}[ht]
    \centering
    \tiny
    \begin{tabular}{c|c|c|c}
    \hline
    Activation Function & Mathematical Form & First Order Derivative Properties & Second Order Derivative Properties \\
    \hline
        ReLU & $g(x)=\max (0, x)$  & $g'(x)=1\cdot I(x>0)+0\cdot I(x<0)$ &  $g''(x)=0  \text { (except at } x=0)$ \\
        Sigmoid & $g(x)=\frac{1}{1+e^{-x}}$ &$\left|g'(x)\right|\leq \frac{1}{4}$ & $\left|g''(x)\right| \leq \frac{1}{4}$ \\
        Tanh & $g(x)=\frac{e^x-e^{-x}}{e^x+e^{-x}}$ & $\left|g'(x)\right|\leq 1$ & $\left|g''(x)\right| \leq 2$ \\
        Leaky ReLU & $g(x)=x\cdot I(x>0) + \alpha x \cdot I(x\leq0)$ & $g'(x)=1\cdot I(x>0) + \alpha\cdot I(x<0)$ &$g''(x)=0  \text { (except at } x=0)$\\
        Swish & $g(x)=x \cdot \frac{1}{1+e^{-x}}$ & $\left|g'(x)\right|<2$ & $\left|g''(x)\right| < 2$ \\
        ELU & $g(x)=x\cdot I(x>0) +\alpha (e^x-1)\cdot I(x\leq0) $& $\left|g'(x)\right|\leq \max(0,\alpha)$ & $\left.g''(x)\leq \left|\alpha\right|  \text { (except at } x=0\right)$ \\
        GELU & $g(x)=x \cdot \Phi(x)$ & $\left|g'(x)\right|\leq 1$ & $\left|g''(x)\right|\leq 0.5$ \\
        Mish & $g(x)=x \cdot \tanh \left(\ln \left(1+e^x\right)\right)$ & $\left|g'(x)\right|\leq 1.5$ & $\left|g''(x)\right|\leq 2$ \\
        Softplus & $g(x)=\ln \left(1+e^x\right)$ & $\left|g'(x)\right|\leq 1$ & $\left|g''(x)\right|\leq \frac{1}{4}$ \\
        
        Softmax & $g_i(x)=\frac{e^{x_i}}{\sum_j e^{x_j}}$ &$ \left|g'_i(x)\right|\leq 1$ & $\left|g''_i(x)\right| \leq 1$ \\
        \hline
    \end{tabular} 
    \caption{Properties of commonly used activation functions and their first and second derivatives.}
    \label{tab:activation}
\end{table*}

\noindent
\textbf{Proof of the inequality.} Considering a neural network $f(\cdot)$ with the following structure:
\begin{equation}
    f(x)=g_k\left(\omega_k\cdots g_1\left(\omega_1 x+b_1\right)+b_2 \cdots+b_k\right),
\end{equation}
where $x$ is the input, $g_i(\cdot)$ is the activation function for the $i$-th layer, $\omega_i$ is the weight matrix, and $b_i$ is the bias vector for the $i$-th layer. Using the chain rule, the first derivative of $f(x)$ can be expressed by:
\begin{equation}
    f^{\prime}(x)=\frac{\partial g_k}{\partial z_k} \cdot \frac{\partial z_k}{\partial x},
\end{equation}
where $z_k=\omega_k g_{k-1}\left(z_{k-1}\right)+b_k.$ The second derivative is given by:
\begin{equation}
    f^{\prime \prime}(x)=\frac{\partial^2 g_k}{\partial z_k^2} \cdot\left(\frac{\partial z_k}{\partial x}\right)^2+\frac{\partial g_k}{\partial z_k} \cdot \frac{\partial^2 z_k}{\partial x^2} .
\end{equation}
The complexity of the derivative mainly comes from the activation functions, as the first derivative of a linear function is constant and its second derivative is zero. Table.~\ref{tab:activation} presents the bounds of the first and second derivatives for basic activation functions widely used in neural networks. For smooth functions, both their first and second derivatives are bounded over the entire domain. For non-smooth activation functions, their first and second derivatives are bounded within the domain where they are defined. Let $\mathbf{R}$ represent the entire space and define the domain of the second derivative of $f$ as follows:
\begin{equation}
    \operatorname{dom}(g'')=\operatorname{dom}(g''_1)\cap\operatorname{dom}(g''_2)\cap\cdots\cap\operatorname{dom}(g''_k).
\end{equation}
Therefore, for any given $z\in\operatorname{dom}(g''),$ we have the following bounds:
\begin{equation}
   \left\|g_i^{\prime}(z)\right\|_2 \leq M_i, \quad\left\|g_i^{\prime \prime}(z)\right\|_2 \leq C_i, \quad\left\|\omega_i\right\|_2 \leq W_i, \
\end{equation}
where $M_i$, $C_i$, $W_i$ $i=1, \ldots, k,$ are positive constants. The first derivative of $z_k$ with respect to $x$ in terms of $z_{k-1}$ can be expressed as:
\begin{equation}
    \frac{\partial z_k}{\partial x}=\omega_k \cdot \operatorname{diag}\left(g_{k-1}^{\prime}\left(z_{k-1}\right)\right) \cdot \frac{\partial z_{k-1}}{\partial x}.
\end{equation}
And the second derivative of $z_k$ with respect to $x$ in terms of $z_{k-1}$ can be expressed as:
\begin{equation}
\begin{aligned}
   \frac{\partial^2 z_k}{\partial x^2}&=\omega_k \cdot\left[\operatorname{diag}\left(g_{k-1}^{\prime \prime}\left(z_{k-1}\right)\right) \cdot\left(\frac{\partial z_{k-1}}{\partial x}\right)^2\right] \\
   & +\omega_k \cdot\left[\operatorname{diag}\left(g_{k-1}^{\prime}\left(z_{k-1}\right)\right) \cdot \frac{\partial^2 z_{k-1}}{\partial x^2}\right] .
\end{aligned}
\end{equation}

 Now we use the recursive method to prove that $f''$ is bounded. For the derivative at the first layer:
\begin{equation}
   \left\|\frac{\partial z_1}{\partial x}\right\|_2=\left\|\omega_1\right\|_2 \leq W_1, \quad \left\|\frac{\partial^2 z_1}{\partial x^2}\right\|_2=0.
\end{equation}
Assume that for $z_{i-1}$, its first and second derivatives are bounded:
\begin{equation}
    \left\|\frac{\partial z_{i-1}}{\partial x}\right\|_2 \leq U_{i-1}, \quad\left\|\frac{\partial^2 z_{i-1}}{\partial x^2}\right\|_2 \leq V_{i-1}.
\end{equation}
Then the first derivative of the $i$-th layer satisfies:
\begin{equation}
\begin{aligned}
    \left\|\frac{\partial z_i}{\partial x}\right\|_2 &\leq\left\|\omega_i\right\|_2 \cdot\left\|g_{i-1}^{\prime}\left(z_{i-1}\right)\right\|_2 \cdot\left\|\frac{\partial z_{i-1}}{\partial x}\right\|_2 \\ & = W_i \cdot M_{i-1} \cdot U_{i-1}.
\end{aligned}
\end{equation}
The second derivative also satisfies:
\begin{equation}
    \begin{aligned}
        \left\|\frac{\partial^2 z_i}{\partial x^2}\right\|_2 &\leq\left\|\omega_i\right\|_2 \cdot\left\|g_{i-1}^{\prime \prime}\left(z_{i-1}\right)\right\|_2 \cdot\left\|\frac{\partial z_{i-1}}{\partial x}\right\|_2^2 \\
        &+\left\|\omega_i\right\|_2 \cdot\left\|g_{i-1}^{\prime}\left(z_{i-1}\right)\right\|_2 \cdot\left\|\frac{\partial^2 z_{i-1}}{\partial x^2}\right\|_2 \\
        &= W_i \cdot\left(C_{i-1} \cdot U_{i-1}^2+M_{i-1} \cdot V_{i-1}\right).
    \end{aligned}
\end{equation}
Let 
\begin{equation}
    \begin{aligned}
        U_i &= W_i \cdot M_{i-1} \cdot U_{i-1}, \\
        V_i &= W_i \cdot\left(C_{i-1} \cdot U_{i-1}^2+M_{i-1} \cdot V_{i-1}\right),
    \end{aligned}
\end{equation}
then we have:
\begin{equation}
\left\|\frac{\partial z_i}{\partial x}\right\|_2 \leq U_i, \quad\left\|\frac{\partial^2 z_i}{\partial x^2}\right\|_2 \leq V_i.
\end{equation}
In this way we have proven that $f''(x)$ is bounded within its domain $\operatorname{dom}(g'')$. 

\noindent
\textbf{Proof that the inequality holds almost everywhere.} Let $\mu(\cdot)$ denote the measure of sets, then we need to prove $\mu(\operatorname{dom}(g''))=\mu(\mathbf{R}).$ Reviewing those non-smooth activation functions, their derivatives are only undefined at specific points, and the set of such points is very sparse in high-dimensional space. Take ReLU as an example. The points where its derivative is not defined occur only when the input is zero or when the input is mapped to zero after a linear transformation. Given that the weights in neural networks are typically close to full rank, the set of inputs where ReLU’s derivative is undefined forms a lower-dimensional manifold, which is sparse in high-dimensional space. Consequently, the measure of this set is zero. Therefore $\mu(\mathcal{R}\setminus\operatorname{dom}(g''_i))=0,$ $i=1,\ldots,k.$ Since
\begin{equation}
    \mathbf{R} \setminus \operatorname{dom}(g'') = \left( \mathbf{R} \setminus \operatorname{dom}(g''_1)\right) \cup \cdots \cup \left(\mathbf{R} \setminus \operatorname{dom}(g''_k) \right),
\end{equation}
then we can obtain that
\begin{equation}
\begin{aligned}
    \mu(\mathbf{R}\setminus\operatorname{dom}(g''))&\leq \mu(\mathbf{R}\setminus\operatorname{dom}(g''_1))+\cdots \\
    &+\mu(\mathbf{R}\setminus\operatorname{dom}(g_k))=0.
\end{aligned}
\end{equation}
Therefore, we can conclude that the region where $f''$ is undefined forms a set of measure zero, and the inequality for $f''$ is satisfied almost everywhere. 
Therefore, the function $s(\cdot)$ used to compute the score of latent variables in this paper also satisfies this condition, meaning that there exists a constant $c$ so that for any $\xi$ in the Gaussian space, we have $\|\nabla_\xi^2s(\xi)\|_2\leq c$ almost everywhere.

\section{Hyper-parameters}

\begin{figure}[t]
  \centering
   \includegraphics[width=0.95\linewidth]{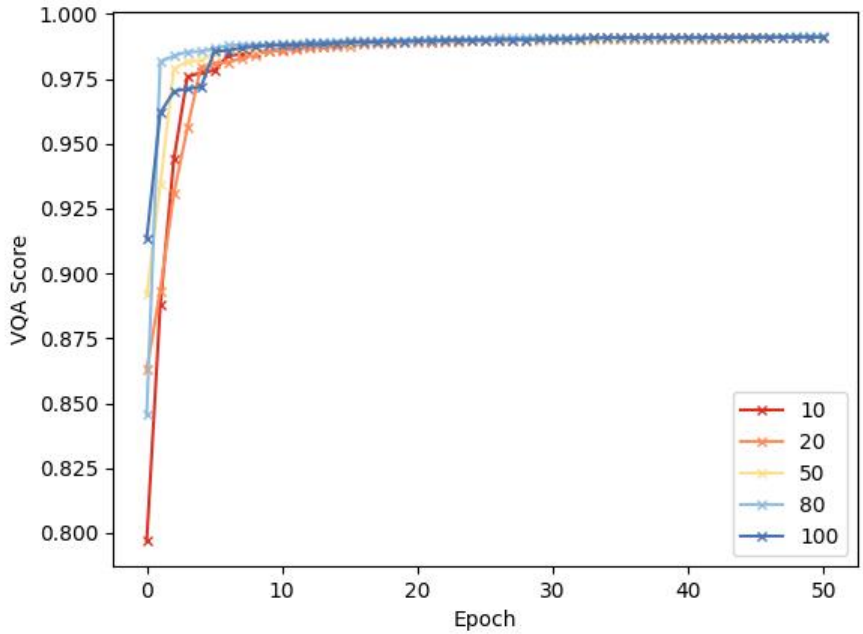}
   \caption{The average VQA score of the generated images across different epochs under various settings of the number of denoising timesteps.}
   \label{fig:timesteps}
\end{figure}

\begin{figure*}[t]
  \centering
   \includegraphics[width=0.95\linewidth]{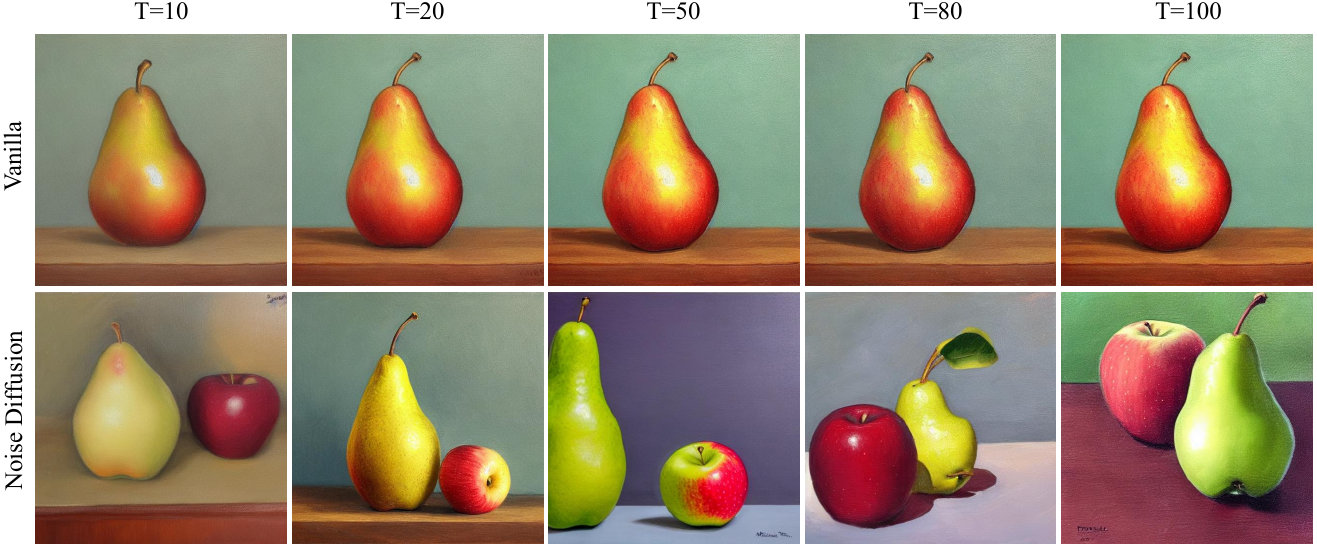}
   \caption{An example of an extreme case. Increasing the number of denoising timesteps does not lead to an improvement in the alignment of the generated image with the prompt. However, under different settings of the number of timesteps, our method consistently enhances the faithfulness.}
   \label{fig:extreme}
\end{figure*}

\begin{figure}[ht]
  \centering
   \includegraphics[width=0.9\linewidth]{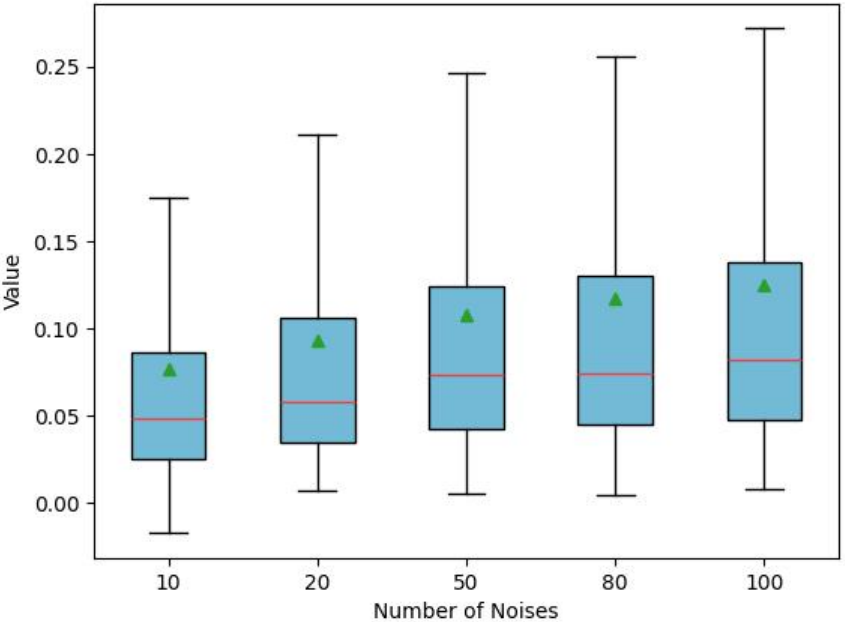}
   \caption{The statistical characteristics of $\nabla_{z_T}s(z_T)v/\|v\|_2^2$ with different values of $N$ during the optimization.}
   \label{fig:noises}
\end{figure}
\subsection{Number oF Denoising Timesteps}
To investigate the impact of the number of denoising timesteps $T$. We select the cases of $T=10,20,50,80,100$ and use the prompt ``an apple and a pear” as an example, conducting the experiment with 25 random seeds. The quantitative results are shown in Fig.~\ref{fig:timesteps}. Regardless of the number of timesteps, our method consistently leads to significant improvements. Generally speaking, increasing the number of timesteps can enhance the VQA score of the initial images; however, the VQA score after increasing the timesteps still leaves room for improvement. Fig.~\ref{fig:extreme} shows an extreme example: increasing $T$ from 10 to 100 does not result in a substantial change in the image content. In contrast, after optimizing the noisy latent variable, the generated images have a significant improvement in alignment with the prompt across different settings. This demonstrates that our method enhances the faithfulness across different numbers of timesteps. When it comes to the choice of $T$, selecting smaller values can significantly speed up inference time while still yielding satisfactory images. If speed is the primary concern, one could opt for $T=20$ or even $T=10$. However, to balance stability with performance, $T=50$ is a recommended choice based on the current performance of diffusion models.

\subsection{Number of Candidate Noises}
The number of candidate noises $N$ should be large enough to ensure that the optimization progresses in the direction of increasing the VQA score. Since the latent variable exists in a high-dimensional space, it is unlikely to find a noise that leads to the step difference perfectly aligning with the gradient through random sampling. Instead, the criterion for setting $N$ is to ensure that the lower bound is sufficiently large. We set $N$ to different values: 10, 20, 50, 80 and 100, and observe the statistical characteristics of $\nabla_{z_T}s(z_T)v/\|v\|_2^2$ under these settings. As shown in Fig.~\ref{fig:noises}, when $N$ increases from 10 to 20, the lower quartile (Q1) improves from 0.026 to 0.035. When $N$ reaches 50, it further increases to 0.043, demonstrating the effect of increasing the number of candidate noises. As $N$ increases to 100, Q1 reaches 0.048. Although the rate of improvement begins to slow down, increasing $N$ further will still lead to continued growth. 
Therefore, although we set $N$ as 50 in our experiments, increasing $N$ undoubtedly has a positive effect on the optimization.

\section{Datasets}
\begin{table*}
    \centering
    \tiny
    \begin{tabular}{cccccc}
        \hline
        an elephant and a rabbit	&	a dog and a frog	&	a bird and a mouse	&	a monkey and a frog	&	a horse and a monkey	&	a bird and a turtle	\\
a bird and a lion	&	a lion and a monkey	&	a horse and a turtle	&	a bird and a monkey	&	a bear and a frog	&	a bear and a turtle	\\
a dog and an elephant	&	a dog and a horse	&	a turtle and a mouse	&	a cat and a turtle	&	a dog and a mouse	&	a cat and an elephant	\\
a cat and a bird	&	a dog and a monkey	&	a lion and a mouse	&	a bear and a lion	&	a bird and an elephant	&	a lion and a turtle	\\
a dog and a bird	&	a bird and a rabbit	&	an elephant and a turtle	&	a lion and an elephant	&	a cat and a rabbit	&	a dog and a bear	\\
a dog and a rabbit	&	a cat and a bear	&	a bird and a horse	&	a rabbit and a mouse	&	a bird and a bear	&	a bear and a monkey	\\
a horse and a frog	&	a cat and a horse	&	a frog and a rabbit	&	a bear and a mouse	&	a monkey and a rabbit	&	a cat and a dog	\\
a lion and a frog	&	a frog and a mouse	&	a dog and a lion	&	a lion and a rabbit	&	an elephant and a frog	&	a frog and a turtle	\\
a cat and a lion	&	a horse and a rabbit	&	a cat and a monkey	&	a bear and a rabbit	&	a turtle and a rabbit	&	an elephant and a monkey	\\
a bird and a frog	&	a lion and a horse	&	a bear and a horse	&	a bear and an elephant	&	a horse and a mouse	&	a dog and a turtle	\\
a monkey and a mouse	&	a cat and a frog	&	a monkey and a turtle	&	a horse and an elephant	&	a cat and a mouse	&	an elephant and a mouse	\\
a horse with a glasses	&	a bear with a glasses	&	a monkey and a red car	&	an elephant with a bow	&	a frog and a purple balloon	&	a mouse with a bow	\\
a bird with a crown	&	a turtle and a yellow bowl	&	a rabbit and a gray chair	&	a dog and a black apple	&	a rabbit and a white bench	&	a lion and a yellow clock	\\
a turtle and a gray backpack	&	an elephant and a green balloon	&	a monkey and a orange apple	&	a lion and a red car	&	a lion with a crown	&	a bird and a purple bench	\\
a rabbit and a orange backpack	&	a rabbit and a orange apple	&	a monkey and a green bowl	&	a frog and a red suitcase	&	a monkey and a green balloon	&	a cat with a glasses	\\
a bear and a blue clock	&	a cat and a gray bench	&	a bear with a crown	&	a lion with a bow	&	a bear and a red balloon	&	a bird and a black backpack	\\
a horse and a pink balloon	&	a turtle and a yellow car	&	a lion with a glasses	&	a cat and a yellow balloon	&	a horse and a yellow clock	&	a dog with a glasses	\\
a horse and a blue backpack	&	a frog with a bow	&	an elephant with a glasses	&	a mouse and a red bench	&	a bird and a brown balloon	&	a monkey and a yellow backpack	\\
a turtle and a pink balloon	&	a cat and a red apple	&	a monkey and a brown bench	&	a rabbit with a glasses	&	a bear and a gray bench	&	a turtle and a blue clock	\\
a monkey and a blue chair	&	a turtle and a blue chair	&	a dog with a bow	&	an elephant and a black chair	&	a mouse and a purple chair	&	a bear and a white car	\\
a lion and a black backpack	&	a dog with a crown	&	a horse and a green apple	&	a dog and a gray clock	&	a dog and a purple car	&	a dog and a gray bowl	\\
a monkey with a bow	&	a mouse and a blue clock	&	a bird and a black bowl	&	a horse and a white car	&	a mouse and a pink apple	&	a bear and a orange backpack	\\
an elephant and a yellow clock	&	a bird and a green chair	&	a mouse and a black balloon	&	a turtle and a white bench	&	a bird with a bow	&	a turtle with a crown	\\
a bird and a yellow car	&	a frog and a orange car	&	a dog and a pink bench	&	a frog with a crown	&	a frog and a green bowl	&	a frog and a pink bench	\\
a horse with a bow	&	a bird and a yellow apple	&	a monkey with a crown	&	a cat and a blue backpack	&	a turtle and a pink apple	&	a dog and a orange chair	\\
a horse and a green suitcase	&	an elephant with a crown	&	a monkey and a orange suitcase	&	a turtle and a orange suitcase	&	a lion and a gray apple	&	a mouse with a crown	\\
a mouse with a glasses	&	a horse and a brown bowl	&	a monkey and a yellow clock	&	a turtle with a bow	&	a dog and a brown backpack	&	a cat and a purple bowl	\\
a lion and a white bench	&	a rabbit and a blue bowl	&	a lion and a brown balloon	&	a horse and a pink chair	&	an elephant and a green bench	&	a rabbit and a white balloon	\\
an elephant and a pink backpack	&	a lion and a orange suitcase	&	an elephant and a orange apple	&	an elephant and a green suitcase	&	a horse with a crown	&	a bear with a bow	\\
a rabbit and a yellow suitcase	&	a horse and a blue bench	&	a dog and a green suitcase	&	a mouse and a red car	&	a cat and a black chair	&	a bear and a red suitcase	\\
a rabbit and a gray clock	&	a bear and a pink apple	&	a lion and a white chair	&	a rabbit with a crown	&	a mouse and a purple bowl	&	a frog and a black apple	\\
a rabbit with a bow	&	a mouse and a pink suitcase	&	a lion and a pink bowl	&	a frog and a black chair	&	a frog and a green clock	&	a bear and a white chair	\\
an elephant and a brown car	&	a turtle with a glasses	&	a cat and a black suitcase	&	a cat and a yellow car	&	a frog and a yellow backpack	&	a bird and a black suitcase	\\
a cat with a crown	&	a rabbit and a yellow car	&	a cat with a bow	&	a bird and a white clock	&	a cat and a green clock	&	a bear and a purple bowl	\\
a monkey with a glasses	&	a frog with a glasses	&	an elephant and a green bowl	&	a bird with a glasses	&	a dog and a blue balloon	&	a mouse and a brown backpack	\\
a pink crown and a purple bow	&	a blue clock and a blue apple	&	a blue balloon and a orange bench	&	a pink crown and a red chair	&	a orange chair and a blue clock	&	a purple bowl and a black bench	\\
a green glasses and a black crown	&	a purple chair and a red bow	&	a yellow glasses and a black car	&	a orange backpack and a purple car	&	a white balloon and a white apple	&	a brown suitcase and a black clock	\\
a yellow backpack and a purple chair	&	a gray backpack and a green clock	&	a blue crown and a red balloon	&	a gray suitcase and a black bowl	&	a brown balloon and a pink car	&	a black backpack and a green bow	\\
a blue balloon and a blue bow	&	a white bow and a white car	&	a orange bowl and a purple apple	&	a brown chair and a white bench	&	a purple crown and a blue suitcase	&	a yellow bow and a orange bench	\\
a yellow glasses and a brown bow	&	a red glasses and a red suitcase	&	a pink bow and a gray apple	&	a gray crown and a white clock	&	a black car and a white clock	&	a brown bowl and a green clock	\\
a green backpack and a yellow crown	&	a orange glasses and a pink clock	&	a purple chair and a orange bowl	&	a orange suitcase and a brown bench	&	a white glasses and a orange balloon	&	a yellow backpack and a gray apple	\\
a green bench and a red apple	&	a gray backpack and a yellow glasses	&	a green glasses and a yellow chair	&	a white glasses and a gray apple	&	a gray suitcase and a brown bow	&	a white car and a black bowl	\\
a purple car and a pink apple	&	a gray crown and a purple apple	&	a orange car and a red bench	&	a red suitcase and a blue apple	&	a red backpack and a yellow bowl	&	a red bench and a yellow clock	\\
a black backpack and a pink balloon	&	a blue suitcase and a gray balloon	&	a yellow glasses and a gray bowl	&	a white suitcase and a white chair	&	a purple crown and a blue bench	&	a yellow bow and a pink bowl	\\
a green backpack and a brown suitcase	&	a green glasses and a black bench	&	a white bow and a black clock	&	a red crown and a black bowl	&	a green chair and a purple car	&	a white chair and a gray balloon	\\
a pink chair and a gray apple	&	a yellow suitcase and a yellow car	&	a green backpack and a purple bench	&	a black crown and a red car	&	a green balloon and a pink bowl	&	a purple balloon and a white clock	\\
        \hline
    \end{tabular}
    \caption{Prompt dataset for simple cases.}
    \label{tab:simple}
\end{table*}

\begin{table*}[ht]
\centering
\tiny
\begin{tabular}{ccccc}
\hline
an elephant under a rabbit	&	a dog under a frog	&	a monkey under a frog	&	a bird on a turtle	&	a bird above a lion	\\
a bird on a monkey	&	a dog on an elephant	&	a cat on an elephant	&	a dog under a monkey	&	a lion under a mouse	\\
a bird above an elephant	&	a lion under a turtle	&	a dog below a bird	&	a bird above a rabbit	&	an elephant under a turtle	\\
a cat under a rabbit	&	a dog on a bear	&	a cat above a bear	&	a bird on a horse	&	a bear under a monkey	\\
a cat on a horse	&	a frog under a mouse	&	a dog on a lion	&	an elephant under a frog	&	a cat on a lion	\\
a horse under a rabbit	&	a cat below a monkey	&	an elephant below a monkey	&	a bird above a frog	&	a cat above a frog	\\
a monkey under a turtle	&	a cat above a mouse	&	an elephant under a mouse	&	a monkey on a red car	&	a frog below a purple balloon	\\
a turtle under a yellow bowl	&	a rabbit on a gray chair	&	a rabbit under a white bench	&	a lion below a yellow clock	&	an elephant below a green balloon	\\
a monkey above a green bowl	&	a bear below a blue clock	&	a bear under a crown	&	a mouse under a red bench	&	a bear under a gray bench	\\
a turtle below a blue clock	&	a monkey on a blue chair	&	a bear below an orange backpack	&	a bird on a green chair	&	a mouse below a black balloon	\\
a turtle on a white bench	&	a bird above a yellow car	&	a monkey under a crown	&	a cat above a blue backpack	&	a dog on an orange chair	\\
a monkey on an orange suitcase	&	a mouse below a crown	&	a horse under a brown bowl	&	a monkey below a yellow clock	&	a lion on a white bench	\\
a rabbit under a blue bowl	&	a dog above a green suitcase	&	a cat under a black chair	&	a lion under a pink bowl	&	a frog on a black chair	\\
a frog on a green clock	&	a pink crown under a purple bow	&	a blue clock under a blue apple	&	a blue balloon on an orange bench	&	a pink crown under a red chair	\\
a purple bowl on a black bench	&	an orange backpack on a purple car	&	a brown suitcase below a black clock	&	a gray backpack under a green clock	&	a brown balloon above a pink car	\\
a purple crown on a blue suitcase	&	a yellow bow above an orange bench	&	a pink bow on a gray apple	&	a gray crown below a white clock	&	a black car under a white clock	\\
a brown bowl below a green clock	&	a green backpack under a yellow crown	&	a purple chair under an orange bowl	&	an orange suitcase under a brown bench	&	a yellow backpack under a gray apple	\\
a green bench under a red apple	&	a gray suitcase under a brown bow	&	a white car under a black bowl	&	a purple car under a pink apple	&	a gray crown under a purple apple	\\
a red suitcase under a blue apple	&	a red backpack under a yellow bowl	&	a black backpack below a pink balloon	&	a blue suitcase below a gray balloon	&	a white suitcase on a white chair	\\
a green backpack under a brown suitcase	&	a white bow on a black clock	&	a red crown under a black bowl	&	a yellow suitcase on a yellow car	&	a green balloon above a pink bowl	\\
\hline
\end{tabular}
\caption{Prompt dataset for complex cases.}
\label{tab:complex}
\end{table*}

We present the prompt datasets used in our main experiment. Table.~\ref{tab:simple} shows the dataset for simple cases, which includes a total of 276 prompts, categorized into three groups: animals, animals-objects, and objects. The complex cases are shown in Table.~\ref{tab:complex}, where we have a total of 100 prompts, which similarly include animals, animals-objects and objects, but with additional spatial relationships such as ``on," ``under," ``above," and ``below." These added relationships increase the complexity of the prompts, forcing text-to-image models to better understand and reflect the logical structure within the prompts.
\section{More Results}
\begin{figure*}[t]
  \centering
   \includegraphics[width=0.95\linewidth]{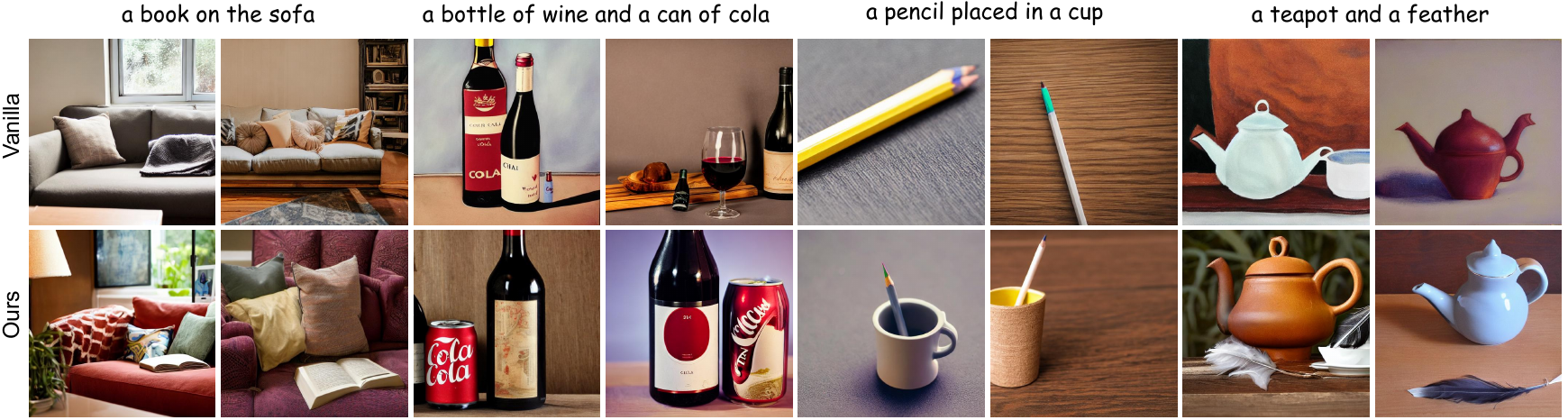}
   \caption{Outputs of Vanilla settings and our method for SD V-1.4.}
   \label{fig:v-1.4}
\end{figure*}

\begin{figure*}[t]
  \centering
   \includegraphics[width=0.95\linewidth]{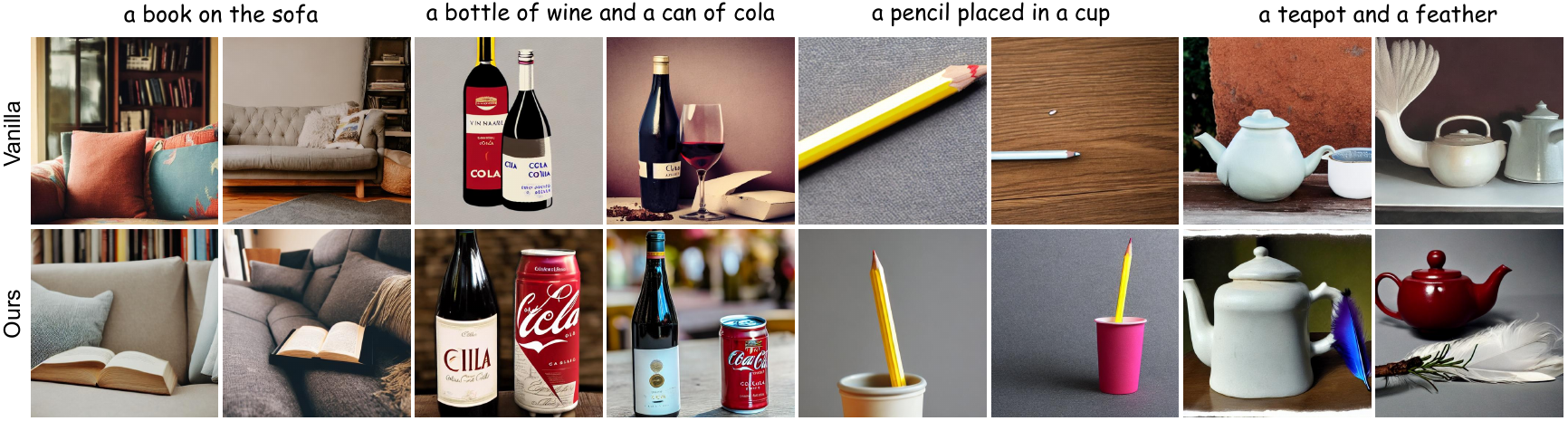}
   \caption{Outputs of Vanilla settings and our method for SD V-1.5.}
   \label{fig:v-1.5}
\end{figure*}

\begin{figure*}[t]
  \centering
   \includegraphics[width=0.95\linewidth]{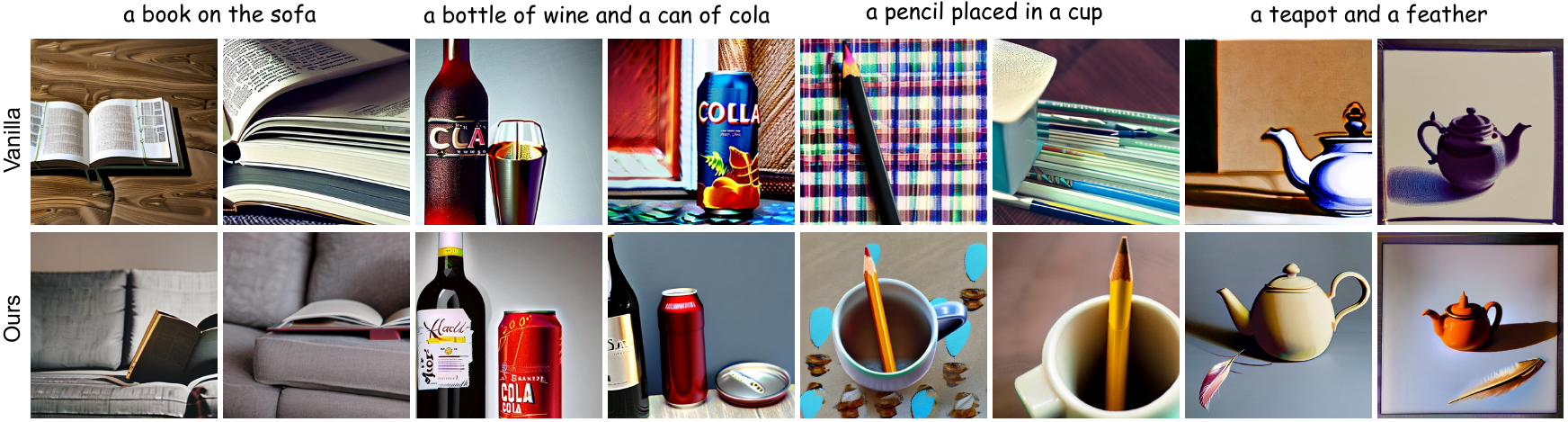}
   \caption{Outputs of Vanilla settings and our method for SD V-2.0.}
   \label{fig:v-2.0}
\end{figure*}

\begin{figure*}[t]
  \centering
   \includegraphics[width=0.95\linewidth]{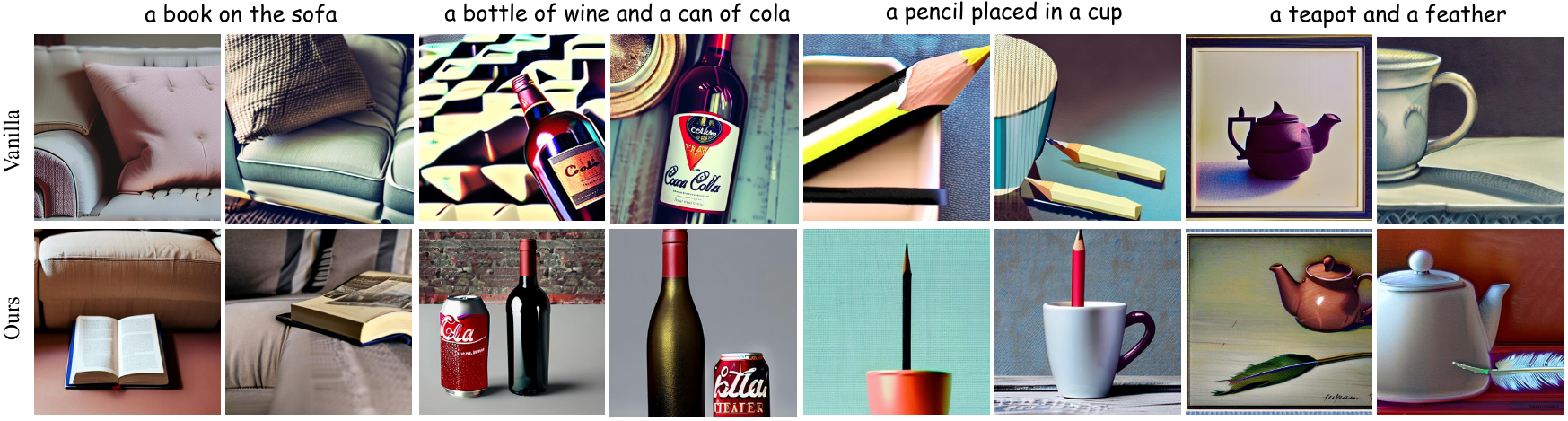}
   \caption{Outputs of Vanilla settings and our method for SD V-2.1.}
   \label{fig:v-2.1}
\end{figure*}

We present additional results for vanilla settings and our method on SD V-1.4, V-1.5, V-2.0 and V-2.1 in Fig.~\ref{fig:v-1.4}, Fig.~\ref{fig:v-1.5}, Fig.~\ref{fig:v-2.0} and Fig.~\ref{fig:v-2.1}, respectively.


\end{document}